\documentclass[sigconf]{acmart}
\usepackage{graphicx}
\usepackage{hyperref}
\usepackage{color}

\usepackage{booktabs}
\usepackage{makecell}
\usepackage{tabularx}
\usepackage{comment}
\usepackage{subcaption}
\usepackage{xcolor}
\usepackage{enumitem}

\usepackage[ruled,vlined]{algorithm2e}

\usepackage{mwe}

\usepackage{algpseudocode}
\usepackage{multirow}
\usepackage{mathtools}
\usepackage{siunitx}
\usepackage{listings}
\usepackage{float} 
\definecolor{mydarkgreen}{rgb}{0.0, 0.5, 0.0}

\AtBeginDocument{%
  }

\copyrightyear{2026}
\acmYear{2026}
\setcopyright{cc}
\setcctype{by}
\acmConference[GECCO Companion '26]{Genetic and Evolutionary Computation Conference}{July 13--17, 2026}{San Jose, Costa Rica}
\acmBooktitle{Genetic and Evolutionary Computation Conference (GECCO Companion '26), July 13--17, 2026, San Jose, Costa Rica}
\acmDOI{10.1145/3795101.3814664}
\acmISBN{979-8-4007-2488-6/2026/07}

\begin{document}

\title{CoupleEvo: Evolving Heuristics for Coupled Optimization Problems Using Large Language Models}


\author{Thomas Bömer}
\email{thomas.boemer@kit.edu}
\orcid{0000-0003-4979-7455}
\affiliation{%
  \institution{Karlsruhe Institute of Technology}
  \city{Karlsruhe}
  \country{Germany}
}

\author{Bastian Amberg}
\email{bastian.amberg@kit.edu}
\orcid{0000-0001-6715-3819}
\affiliation{%
  \institution{Karlsruhe Institute of Technology}
  \city{Karlsruhe}
  \country{Germany}
}

\author{Max Disselnmeyer}
\email{max.disselnmeyer@kit.edu}
\orcid{0009-0008-5689-2235}
\affiliation{%
  \institution{Karlsruhe Institute of Technology}
  \city{Karlsruhe}
  \country{Germany}
}

\author{Anne Meyer}
\email{anne.meyer@kit.edu}
\orcid{0000-0001-6380-1348}
\affiliation{%
  \institution{Karlsruhe Institute of Technology}
  \city{Karlsruhe}
  \country{Germany}
}

\renewcommand{\shortauthors}{Bömer et al.}

\begin{abstract}
Many real-world optimization problems consist of multiple tightly coupled subproblems whose solutions must be coordinated to achieve high overall performance. 
However, existing large language model driven automated heuristic design approaches are limited to single-problem settings.
In this paper, we propose \texttt{CoupleEvo}.
\texttt{CoupleEvo} proposes three evolutionary coordination strategies to evolve heuristics for coupled optimization problems: the sequential strategy evolves heuristics for one subproblem after the other; the iterative strategy alternates the evolution of heuristics for different subproblems over successive generations; and the integrated strategy evolves heuristics for all problems simultaneously.
The approach is evaluated on two representative coupled optimization problems.
Experimental results show that decomposition-based strategies (sequential and iterative) provide more stable convergence and higher solution quality, while the integrated evolution strategy suffers from increased search complexity and variability.
These findings highlight the importance of coordinating evolutionary search across interdependent subproblems and demonstrate the potential of LLM-driven heuristic design for complex coupled optimization problems.
The code is available: \url{https://github.com/tb-git-kit-research/CoupleEvo}.
\end{abstract}


\begin{teaserfigure}
\centering
  \includegraphics[width=\textwidth]{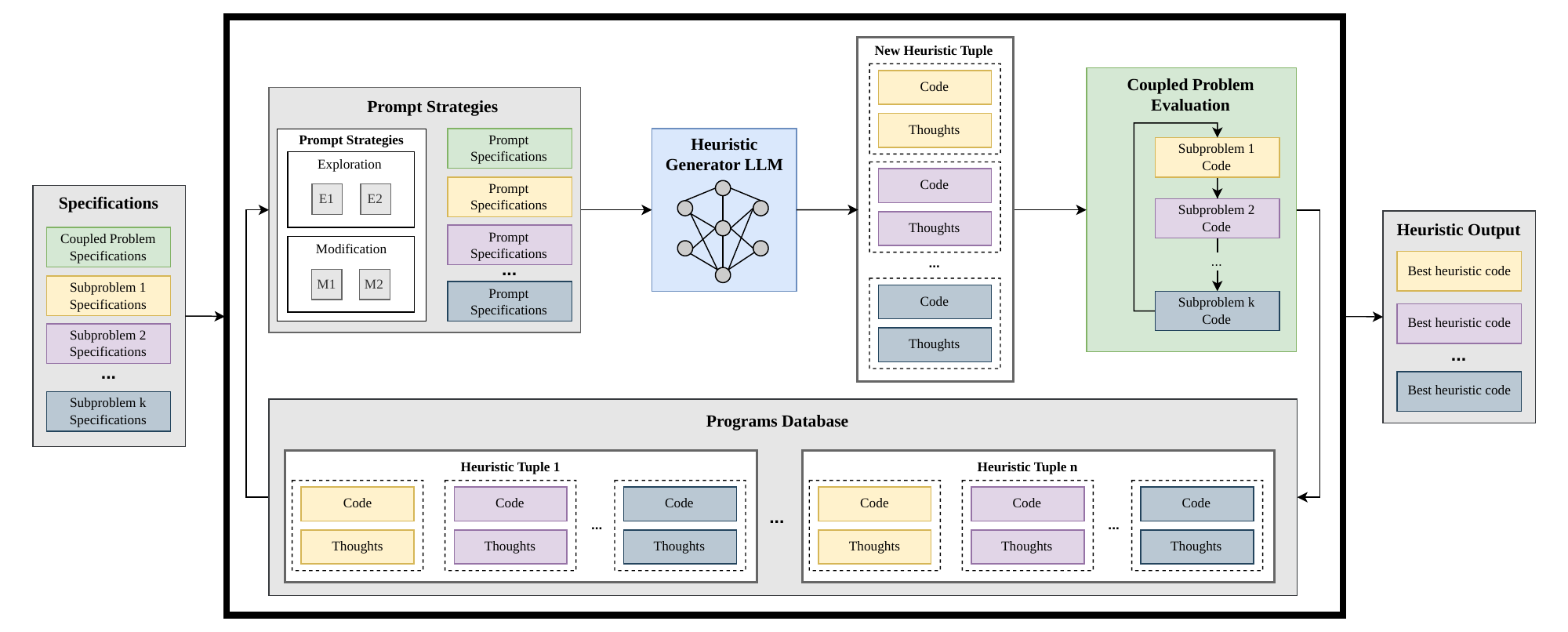}
  \caption{CoupleEvo evolves multiple heuristics for coupled optimization problems.}
  \label{fig:CoupleEvo framework concept overview}
\end{teaserfigure}


\maketitle

\section{Introduction}
\label{sec:Introduction}

\begin{figure}[!hptb]
    \centering
    \begin{subfigure}[t]{0.25\linewidth}
        \centering
        \includegraphics[width=\linewidth]{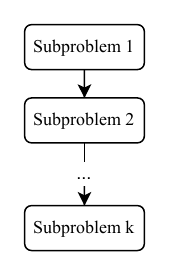}
        \caption{sequential}
        \label{fig:sequential_solution}
    \end{subfigure}\hfill
    \begin{subfigure}[t]{0.25\linewidth}
        \centering
        \includegraphics[width=\linewidth]{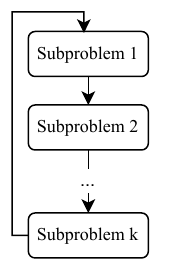}
        \caption{iterative}
        \label{fig:iterative_solution}
    \end{subfigure}\hfill
    \begin{subfigure}[t]{0.25\linewidth}
        \centering
        \includegraphics[width=\linewidth]{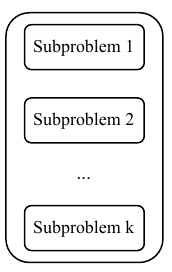}
        \caption{integrated}
        \label{fig:integrated_solution}
    \end{subfigure}
    \caption{Comparison of sequential, iterative, and integrated solution strategies for coupled optimization problems.}
    \label{fig:solution_strategies}
\end{figure}

Many real-world optimization systems are composed of multiple tightly coupled subproblems whose solutions must be jointly coordinated in order to achieve high overall system performance. 
These \emph{coupled optimization problems} often combine two or more NP-hard optimization subproblems.
This structure is prevalent, for example, in complex production and logistics environments.
In this domain, coupled optimization problems frequently include a vehicle routing component. Representative examples of such coupled problems include the Inventory Routing Problem (IRP) \cite{campbell1998inventory}, the Production Routing Problem \cite{ADULYASAK2015141},  and the Location Routing Problem \cite{PRODHON20141}.
In these settings, the interaction among subproblems fundamentally determines solution quality. 
Consequently, optimizing each subproblem in isolation often yields suboptimal global outcomes. 
At the same time, the combinatorial growth of the joint solution space renders fully integrated, simultaneous optimization computationally intractable for realistically sized instances. 
As a result, practical solution approaches commonly rely on some form of sequential or iterative coordination between subproblems, rather than integrated global optimization (see Figure \ref{fig:solution_strategies}).
\par

In recent work, large language models (LLMs) have demonstrated strong performance in the automatic design of heuristics for combinatorial optimization problems \cite{liu2025seoh,romera2024mathematical,van2024llamea,ye2024reevo}.
In these LLM-based automated heuristic design frameworks the evolutionary loop evolves a single heuristic for a single-problem setting.
The challenge to design multiple heuristics for multiple tightly coupled subproblems has not been addressed yet. 
For the adoption of LLM-enabled heuristic design, it requires an \textit{evolutionary coordination strategy} to coordinate the evolution of multiple heuristics.
In this context, the evolutionary coordination strategy itself becomes a first-class design object that directly shapes the evolutionary search dynamics and the resulting quality of generated heuristics. 
This observation motivates the systematic investigation of evolutionary coordination strategies for coupled problems in LLM-driven automated heuristic design.
\par
To address this gap, we propose \textbf{\texttt{CoupleEvo}} (see Figure \ref{fig:CoupleEvo framework concept overview}), which comprises three evolutionary coordination strategies inspired by established decomposition-based  approaches for LLM-driven automated heuristic design in coupled optimization problems:

\begin{itemize}
\item \textbf{Sequential Evolution:} Heuristics are evolved for one subproblem after another.
\item \textbf{Iterative Evolution:} Heuristics are evolved alternating between subproblems over multiple rounds.
\item \textbf{Integrated Evolution:} Heuristics for all subproblems are evolved simultaneously.
\end{itemize}

Each evolutionary coordination strategy has its own potential advantages and disadvantages.
The sequential evolution strategy offers conceptual simplicity and a strong local convergence. 
However, premature convergence in the evolution of an earlier-stage subproblem may constrain the effective search space for subsequent subproblems. 
This may impede the discovery of globally high-quality and well-coordinated heuristics.
The iterative evolution strategy seeks to address this limitation by repeatedly shifting the evolutionary focus between subproblems. This enables progressive refinement of each component while preserving flexibility for coordination.
At the same time, the frequent alternation between subproblems may prevent the evolutionary process from fully exploiting the local optima of each subproblem, thereby limiting the overall potential of the search.
By contrast, the integrated evolution strategy models the interdependence between subproblems directly by evolving all subproblem heuristics concurrently. 
This enables the evolutionary process to explicitly capture and exploit cross-component interactions. 
While this approach offers the greatest potential for discovering highly coordinated, high-quality solutions, it comes at the cost of increased heuristic search complexity and substantially higher demands on the generative capabilities of the LLM.
\par
In this work, we investigate the evolutionary coordination strategies for coupled optimization problems within an iterative Large Neighborhood Search (LNS) framework \cite{ropke2006adaptive}. Each subproblem is modeled as an independent LNS component, and the overall solution is obtained by iteratively coordinating the LNS components of all subproblems.
LLMs are employed as automated heuristic designers to generate a dedicated destroy heuristic for each subproblem. The destroy heuristic determines which components of the current solution of its corresponding subproblem are removed in the iteration. 
The repair heuristic stays deterministic.
Through this mechanism, the LLMs directly control neighborhood exploration and guide the search process.
\par
\par

The main contributions of this work are summarized as follows:
\begin{itemize}
    \item \textbf{\texttt{CoupleEvo}:} We formulate the first LLM-driven framework for automated heuristic design in coupled optimization problems, extending existing single-problem approaches to settings with multiple interdependent NP-hard subproblems.
    
    \item \textbf{Evolutionary Strategies:} We propose three evolutionary coordination strategies (\emph{Sequential Evolution}, \emph{Iterative Evolution}, and \emph{Integrated Evolution}) that govern how heuristics for interacting subproblems are generated within the evolutionary process.
    
    \item \textbf{Comprehensive Evaluation:} We conduct extensive experimental evaluations on two 
    representative coupled optimization domains 
    (Inventory Routing Problem
    and Multi-robot Multi-bay Unit-Load Pre-Marshalling Problem) 
    providing empirical insights into the effectiveness and trade-offs of different evolutionary coordination strategies.
\end{itemize}

\par
The rest of the paper is organized as follows. 
Section \ref{sec: coupled optimization problems} briefly introduces the 
two
studied combinatorial optimization problems. 
Section \ref{sec: related work} provides an overview on automated heuristic design approaches using LLMs.
Section \ref{sec:CoupleEVO: Evolutionary Coordination Strategies for Coupled Optimization Problems} presents the evolutionary procedure and the proposed evolutionary coordination strategies. 
Section \ref{sec: Computational Experiments} presents the results of computational experiments. 
Finally, Section \ref{sec: Conclusion} summarizes the findings and outlines future research directions.

\section{Coupled Optimization Problems}
\label{sec: coupled optimization problems}
To analyze the evolutionary strategies proposed in \texttt{CoupleEvo}, we consider two representative coupled optimization problems. 
The Inventory Routing Problem represents the class of well-established, canonical benchmark problems in coupled optimization. 
In contrast, the Multi-robot Multi-bay Unit-load Pre-marshalling Problem represents a recently emerging coupled optimization problem that is less commonly covered in existing literature and, consequently, less likely to be well represented in LLM knowledge bases.

\paragraph{Inventory Routing Problem.}
The Inventory Routing Problem (IRP) \cite{campbell1998inventory} is a classical coupled optimization problem that integrates inventory management and vehicle routing decisions.
The objective is to determine replenishment quantities and delivery routes for a set of customers over a planning horizon, such that customer demand is satisfied while minimizing total cost, including transportation and inventory holding costs.
\par
The problem can be decomposed into two tightly coupled subproblems: the \textit{inventory management problem} and the \textit{capacitated vehicle routing problem}.
The inventory management problem involves determining the amount and time to deliver inventory to each customer in order to maintain feasible stock levels over time.
Given these replenishment decisions, the vehicle routing problem constructs delivery routes that satisfy vehicle capacity constraints and minimize transportation cost.


\paragraph{Multi-robot Multi-bay Unit-load Pre-marshalling Problem.}
The Multi-robot Multi-bay Unit-load Pre-marshalling Problem (MR-MUPMP) \cite{BOMER2026107359sorting,bomer2024sorting,BOMER2025508CPsort} generalizes the Unit-load Pre-marshalling Problem \cite{pfrommer2023solving} into a tightly coupled optimization setting.
The objective is to reorganize unit loads in a block-stacking warehouse according to priority classes using a fleet of autonomous robots, while minimizing the overall makespan of the operations.
\par
The problem can be decomposed into two interdependent subproblems: the \textit{move search problem} and a \textit{vehicle routing problem with precedences}.
The move search problem determines a sequence of relocation operations required to reach a blockage-free configuration, where each move consists of repositioning a unit load from one storage slot to another.
The vehicle routing problem with precedences then assigns these moves to robots and schedules their execution so as to minimize the total makespan.
These precedences arise from the fact that certain moves must be performed before others in order to ensure executability.

\section{Related Work}
\label{sec: related work}
In this section, we review studies that employ LLMs within evolutionary frameworks for heuristic design in combinatorial optimization. For a broader perspective, \cite{liuLITreviewLLM} provides a comprehensive survey of LLMs for algorithm design, including a clear taxonomy and an extensive overview of existing approaches.
\par
LLM-based automated heuristic design was pioneered by Google DeepMind’s \textbf{FunSearch} \cite{romera2024mathematical}, which leverages a pre-trained LLM within an evolutionary search loop to iteratively generate and refine a single heuristic code. A variety of subsequent approaches have since extended and adapted this paradigm.
\textbf{EoH} \cite{liu2024evolutionEoH} represents heuristics as natural-language thoughts and executable code, refining both jointly through structured prompting. It achieves superior performance with significantly fewer LLM queries than prior approaches.
\textbf{P-CEoH} \cite{bomer2025leveraging} and \textbf{A-CEoH} \cite{bomer2026algorithmic} enhance EoH through prompt augmentation, showing that in-context learning improves heuristic performance and evolutionary efficiency, particularly for smaller LLMs.
\textbf{ReEvo} \cite{ye2024reevo} augments evolutionary search with reflective reasoning through a multi-stage workflow, using iterative reflection to guide heuristic generation and improve performance.
\textbf{HSEvo} \cite{dat2024hsevo} combines evolutionary operators with harmony search and a reflection mechanism to enhance diversity and tune elitist heuristic parameters.
\textbf{Hercules} \cite{2024efficientHERCULES} improves prompt efficiency through reflective abstraction of elite heuristics and reduces evaluation cost via performance prediction based on semantic similarity.
\textbf{HiFo-Prompt} \cite{chen2025hifo} augments LLM-based heuristic evolution with foresight and hindsight mechanisms that adapt search dynamics and distill reusable knowledge to guide future generations.
\textbf{ShinkaEvolve} \cite{lange2025shinkaevolve} introduces a evolutionary framework that improves sample efficiency in LLM-driven program evolution through adaptive parent selection, novelty-based rejection sampling, and bandit-based LLM ensemble selection.
\textbf{EvoX} \cite{liu2026evox} introduces a meta-evolutionary framework that jointly evolves heuristics and their search strategies. It adapts selection and variation dynamically to balance exploration and exploitation.
\textbf{EASE} \cite{viktorin2025solve} provides a modular and domain-agnostic framework for LLM-guided algorithm design. It enables flexible orchestration of components, supporting modular research across diverse tasks.
\textbf{LLaMEA} \cite{van2024llamea} applies LLM-guided evolutionary design to continuous black-box optimization, generating and refining full metaheuristic algorithms that outperform state-of-the-art methods.
\par
Recent work has extended automated heuristic design to the generation of LNS components. Notably, some LNS-focused approaches move beyond single-heuristic design and consider the joint generation of two interacting heuristics.  
\textbf{LLM-LNS} \cite{ye2025largeLLMLNS} applies LLM-driven heuristic evolution to the design of a single destroy operator within an LNS framework for Mixed Integer Linear Programming.  
\textbf{G-LNS} \cite{zhao2026glns} evolves two heuristics—a destroy and a repair operator—within a cooperative evolutionary framework. The operators are maintained in separate populations and evaluated jointly by sampling pairs during LNS execution, while a synergy matrix captures their interaction and guides selection. The framework is evaluated on two standard routing problems.  
\textbf{VRP-Agent} \cite{hottung2025vrpagent} proposes an LLM-driven framework that evolves two complementary heuristics, namely a destroy operator and an ordering operator, for LNS. In contrast to separate populations, heuristic components are generated and maintained as pairs. The repair operator remains the same. The approach is evaluated on routing problems. 
\par
Overall, existing work demonstrates that LLMs can effectively generate heuristics for combinatorial optimization. Most works focus on framework extensions to improve performance through refined evolutionary mechanisms. 
The adaptation to LNS has further motivated the transition from single to multiple interacting heuristics. 
However, to the best of our knowledge, no prior work addresses the generation of heuristic tuples for coupled optimization problems, nor systematically studies different evolutionary coordination strategies.

\section{CoupleEvo: Evolutionary Coordination Strategies for Coupled Optimization Problems}
\label{sec:CoupleEVO: Evolutionary Coordination Strategies for Coupled Optimization Problems}
In this section, we describe the general evolutionary loop and the sequential, iterative, and integrated evolutionary coordinating strategies proposed in \texttt{CoupleEvo}. 

\paragraph{Evolutionary Procedure.}
We consider a coupled optimization problem composed of a set of subproblems $\mathcal{K} = \{1,\dots,|\mathcal{K}|\}$. 
We define a \textit{heuristic tuple} as:
\[
    h = \big( h^{(1)}, h^{(2)}, \dots, h^{(|\mathcal{K}|)} \big),
\]
where $h^{(k)}$ denotes the heuristic associated with subproblem $k \in \mathcal{K}$. 

Each component heuristic $h^{(k)}$ is represented as a pair
\[
    h^{(k)} = \big( c^{(k)}, \tau^{(k)} \big),
\]
where $c^{(k)}$ denotes the executable \emph{program code} implementing the heuristic logic and $\tau^{(k)}$ denotes the corresponding \emph{thought} describing the underlying conceptual idea that guided the construction of $c^{(k)}$.

Accordingly, a full heuristic tuple takes the form:
\[
    h = \big( (c^{(1)}, \tau^{(1)}), \dots, (c^{(|\mathcal{K}|)}, \tau^{(|\mathcal{K}|)}) \big).
\]

\par
The evolutionary framework refines heuristics for these subproblems over multiple generations.
The evolutionary loop differs for the three evolutionary coordination strategies: sequential, iterative, and integrated.
\par
The \textbf{sequential} strategy progresses through the set of subproblems $\mathcal{K}$.
For each subproblem $k \in \mathcal{K}$, an independent evolutionary procedure is executed for a predefined number of generations $\bar{g}$.
The population of heuristic tuples at generation $g \in \mathcal{G}$ for subproblem $k \in \mathcal{K}$ is denoted by $\mathcal{P}_{g}^{k}$.
During the evolution of subproblem $k$, only the heuristic $h^{(k)}$ is requested via prompt to evolve, while the remaining heuristic $\{h^{(j)} : j \neq k\}$ are fixed to their current best-performing heuristics and are copied unchanged into every individual of the population.
To evolve subproblem $k$ in generation $g$ a set of prompt strategies $\mathcal{S} = \{1, 2, \dots, |\mathcal{S}|\}$ is employed to produce the next population $\mathcal{P}_{g}^{k}$ using a pre-trained LLM. 
Every strategy $s \in \mathcal{S}$ is applied $\bar{r}$ times, yielding a total of $|\mathcal{S}| \cdot \bar{r}$ new heuristics for generation $g$.
Each newly produced heuristic tuple $h$ is evaluated on a collection of coupled optimization problem instances $\mathcal{I} = \{1, 2, \dots, |\mathcal{I}|\}$, where its performance is quantified by a fitness score $f^{\mathcal{I}}(h)$.
Fitness evaluation is therefore always performed for full heuristic tuples, but evolution is restricted to the currently active subproblem. 
All subproblems that have not been evolved yet are initialized using a simple seed heuristic to allow the fitness evaluation.
After evaluation, $h$ is added to the current population $\mathcal{P}_{g}^{k}$, and the process continues until all heuristics for the generation have been generated.
The population size is denoted by $\bar{n}$.
At the end of each generation, the top-performing $\bar{n}$ heuristics are retained and carried over to the next generation. 
After completing $\bar{g}$ generations, the best-performing heuristic $h^{(k)}$ yet is fixed and used in the evaluation of all subsequent subproblems.
Before the loop begins for subproblem $k$, an initialization phase generates the initial population $\mathcal{P}_{0}^{k}$ by requesting an initial prompt $2 \cdot \bar{r}$ times.
This procedure is summarized in Algorithm~\ref{alg:sequential}.

\begin{algorithm}[htpb]
\caption{Sequential Evolution of Heuristic Tuples}
\label{alg:sequential}

Initialize seed heuristic tuple
\[
h^\star \leftarrow (h^{(1)}_{\text{seed}},\dots,h^{(K)}_{\text{seed}})
\]

\For{$k \in \mathcal{K}$}{
    Generate $\mathcal{P}_0^k$ by requesting an initial prompt $2\bar r$ times \\
    Select the top $\bar n$ individuals from $\mathcal{P}_0^k$ and copy them to $\mathcal{P}_1^k$ \\

    \For{$g = 1$ \KwTo $\bar{g}$}{
        \For{$s \in \mathcal{S}$}{
            \For{$r = 1$ \KwTo $\bar r$}{

                Generate candidate heuristic $h^{(k)}$ using prompt strategy $s$ via the LLM

                Construct full heuristic tuple
                \[
                h = (h^{(1)},\dots,h^{(k)},\dots,h^{(K)})
                \]
                using fixed $h^{(j)}$ for all $j \neq k$

                Evaluate fitness $f^{\mathcal{I}}(h)$ on instance set $\mathcal{I}$

                Insert $h$ into $\mathcal{P}_g^k$
            }
        }

        Select the top $\bar n$ individuals from $\mathcal{P}_g^k$ and copy them to $\mathcal{P}_{g+1}^k$
    }

    Let $h^{(k)}_{\text{best}}$ be the best individual in $\mathcal{P}_{\bar{g}}^k$

    Update
    \[
    h^\star \leftarrow (h^{(1)},\dots,h^{(k)}_{\text{best}},\dots,h^{(K)})
    \]
}

\Return $h^\star$
\end{algorithm}

\par
The \textbf{iterative} evolutionary coordination strategy works similarly to the sequential strategy but employs an additional \emph{round-based} control structure. 
Let $\bar{l}$ denote the number of control rounds.
Let $\mathcal{L} = \{1,2,\dots,\bar{l}\}$ denote the set of control rounds.
At each round $l \in \mathcal{L}$, the algorithm cycles through the set of subproblems $\mathcal{K}$ as described for the sequential strategy. 
The initialization procedure is only performed in the first round $l = 1$ for each subproblem.
In all consecutive rounds, the top $\bar n$ best-performing component heuristics obtained in the previous round for subproblem k are retained.
The sequential strategy can be described as the iterative strategy with only a single control round $l$.
\par
The \textbf{integrated} strategy evolves \emph{all} subproblem heuristics of a heuristic tuple $h$ \emph{simultaneously}. 
The evolutionary loop is similar to the sequential loop described in Algorithm \ref{alg:sequential}, with the difference that the LLM is prompted to produce a new candidate heuristic tuple in its entirety. It jointly proposes updated program code and corresponding thoughts for all subproblems within a single prompt.
The LLM generates a full heuristic tuple. The tuple is evaluated for the coupled optimization problem. 


\paragraph{Heuristic Prompt Strategies}
This work employs the prompt strategies by \cite{liu2024evolutionEoH}, including the initialization prompt I1, two exploration prompts (E1 and E2), and two modification prompts (M1 and M2). 
The following summarizes the functionality of each strategy.

\begin{itemize}[label={}]
  \item \textbf{I1:} Construct a heuristic designed to address the specified optimization problem.
  \item \textbf{E1:} Create a heuristic that is significantly different from $\bar{p}$ parent heuristics sampled from the current population.
  \item \textbf{E2:} Generate a heuristic that preserves the main conceptual principles of $\bar{p}$ parent heuristics while introducing a novel variation.
  \item \textbf{M1:} Refine an existing parent heuristic from the population to improve its effectiveness.
  \item \textbf{M2:} Adjust the parameter settings of a parent heuristic from the population to enhance its overall performance.
\end{itemize}
Please visit our repository for detailed information on the prompts \href{https://github.com/tb-git-kit-research/CoupleEvo/blob/main/docs/prompt_overview.md}{(click here)}.  

\begin{figure}[!hptb]
    \centering
    \includegraphics[width=\linewidth]{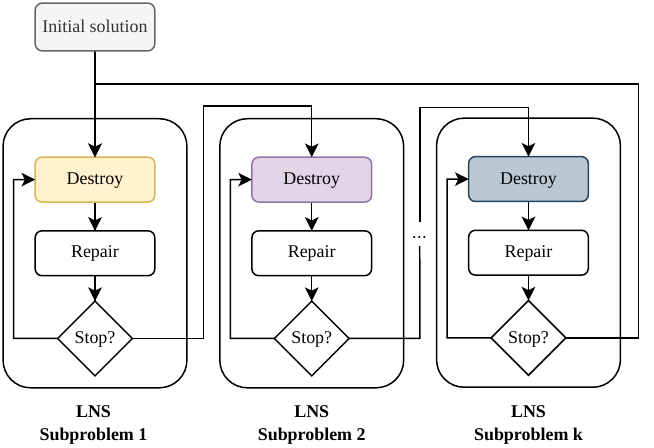}
    \caption{The LLM generates a destroy function for an alternating LNS framework. The LLM-generated parts are colored.}
    \label{fig:alternating_lns}
\end{figure}

\paragraph{Experimental Setup}
\texttt{CoupleEvo} is designed problem-agnostic and can be applied to the automated generation of interacting heuristics in various solution construction or improvement frameworks. 
In this work, the LLM-generated heuristics are deployed as \emph{destroy operators} within an alternating Large Neighborhood Search (LNS) framework for coupled optimization problems \cite{ropke2006adaptive}.
Each subproblem is assigned a dedicated LNS module consisting of a problem-specific destroy and repair operator.
The destroy operator is subject to the LLM-enabled automated heuristic design.
The repair operators are deterministic and identical across all experiments, ensuring that performance differences arise solely from the evolved destroy heuristics.
Figure~\ref{fig:alternating_lns} illustrates the solution procedure and highlights the components evolved by the LLM.

\par
Starting from a simple initial feasible solution, the solution procedure iterates the LNS components of the subproblems 
until a global time limit $T_{\max}$ is reached.
Within each subproblem, the corresponding LNS inner loop terminates when one of its local stopping criteria is met, namely reaching a maximum number of iterations $I_{\max}^{\text{sub}}$ or exceeding a time limit $T_{\max}^{\text{sub}}$.
Upon termination, only the best solution found by the active subproblem LNS is retained and passed as the initial state to the next subproblem’s LNS module.
\par
For each LNS invocation, the corresponding LLM-generated destroy heuristic determines which elements of the current solution are removed.
The repair operator then reconstructs a complete feasible solution from the partially destroyed state.
All feasibility constraints across subproblems are enforced, guaranteeing that intermediate and final solutions remain valid.
\par
This alternating LNS structure is highly modular and readily adaptable to coupled optimization problems across different application domains.
New problem settings can be addressed by defining additional subproblem-specific LNS modules, without modifying the overall evolutionary coordination strategy.

\section{Computational Experiments}
\label{sec: Computational Experiments}
We conducted extensive experiments on 
two
coupled optimization problems to show the effects of the proposed sequential, iterative, and integrated evolutionary coordination strategy.

\paragraph{Evolutionary Parameters.}
We select the \texttt{Qwen3-coder:30b} open-source LLM. The LLM has demonstrated strong coding capabilities \cite{qwen3-coder_30b}.
Each evolutionary coordination strategy (sequential, iterative, and integrated) is executed in six experimental runs 
for each coupled optimization problem 
(IRP 
and MR-MUPMP).
Each of the coupled optimization problems consists of two subproblems.
The population size is set to $\bar{n} = 20$ for all evolutionary coordination strategies.
For the sequential strategy we set the number of generations per subproblem $\bar{g} = 20$ for both subproblems ($20 \cdot 2 = 40$ generations total). 
For the iterative strategy we set the number of control rounds $\bar{l} = 4$ and the number of generations per subproblem $\bar{g} = 5$ ($4 \cdot 2 \cdot 5 = 40$ generations total).
For the integrated strategy we set the number of generations for the coupled problem $\bar{g} = 40$.
Hence, overall we end up with 40 generations total for all evolutionary coordination strategies.
For all evolutionary coordination strategies we employ the prompt strategies E1, E2, M1, and M2. 
The number of parent heuristics for E1 and E2 is set to $\bar{p}=5$. 
The number of prompts per prompt strategy is set to $\bar{r} =20$.
The initialization prompt I1 is called $40$ times for each subproblem. 
In total, we request $40 \cdot 2 + 40 \cdot 4 \cdot 20  = 3280$ prompt per experimental run for all evolutionary coordination strategies. 
This fixed evolutionary budget allows a fair comparison of performance across all evolutionary coordination strategies.

\paragraph{Evaluation Parameters.}
For the 
IRP 
we set the global time limit $T_{\max} = 60$ seconds.
Within each subproblem, the LNS procedure is executed with identical local stopping parameters. 
Specifically, the maximum number of LNS iterations for both subproblems is set to $I_{\text{max}}^{\text{LNS}} = 1000$, and the runtime limit is set to $T_{\text{max}}^{\text{LNS}} = 10$ seconds. 
For the MR-MUPMP we set the global time limit $T_{\max} = 60$.
The maximum number of LNS iterations for the move search is set to $I_{\text{max}}^{\text{MS-LNS}} = 10$ and the subproblem-specific runtime limit is set to $T_{\text{max}}^{\text{MS-LNS}} = 10$ seconds. 
The vehicle routing problem with dependencies is limited to $I_{\text{max}}^{\text{VRP-LNS}} = 90$ iterations per LNS and a runtime of a maximum of $T_{\text{max}}^{\text{MS-LNS}} = 10$ seconds.
Please note, that the heuristic performance is also subject to this parameter setting. 
However, for this work parameter optimization is out of scope. 
\par
For the IRP we use the three instances published by the authors of \cite{UCHOA2017845IRP} with 25 nodes, 18 unit capacity, clustered node positioning, and urban area setting (available here: \url{http://www.iot.ntnu.no/axiom/}).
For the MR-MUPMP we employ ten instances (seed 0-9) published in \cite{BOMER2026107359sorting} with warehouse layout 2x2, bay layout 5x5, fill level 80\,\%, max priority class five, access direction south and north, and robot number two (available here: \url{https://zenodo.org/records/11093870}).  
All heuristics were evaluated on a machine with an AMD EPYC 7401P processor and 64 gigabytes of RAM.

\paragraph{Fitness Calculation}
Each complete heuristic tuple $h$ is evaluated for its fitness. A lower fitness value indicates a better performance.
Let $\mathcal{I}$ be the set of problem instances, $o_i$ the objective value of heuristic tuple $h$ to solve instance $i \in \mathcal{I}$. 
For the IRP, we calculate the fitness as the relative gap between the objective value $o_i$ and the current state-of-the-art HGS approach's objective value $o^{ref}_i$ by \cite{ZHAO2025HGS} (see Equation \ref{eq:fitness_funtion_rel_gap}).

\begin{equation}
\label{eq:fitness_funtion_rel_gap}
f^{\mathcal{I}}(h) = 
\frac{1}{|\mathcal{I}|}
\displaystyle\sum_{i \in \mathcal{I}}
\frac{o_{i} - o^{ref}_i}{o^{ref}_i}
\end{equation}

For the newer MR-MUPMP, we calculate the fitness as the average objective value (see Equation \ref{eq:fitness_funtion_objective}).

\begin{equation}
\label{eq:fitness_funtion_objective}
f^{\mathcal{I}}(h) = 
\frac
{
1
}
{
|\mathcal{I}|
}
\displaystyle\sum_{i \in \mathcal{I}}{o_{i}}
\end{equation}

\begin{figure}[hptb]
    \centering
    \includegraphics[width=1\linewidth]{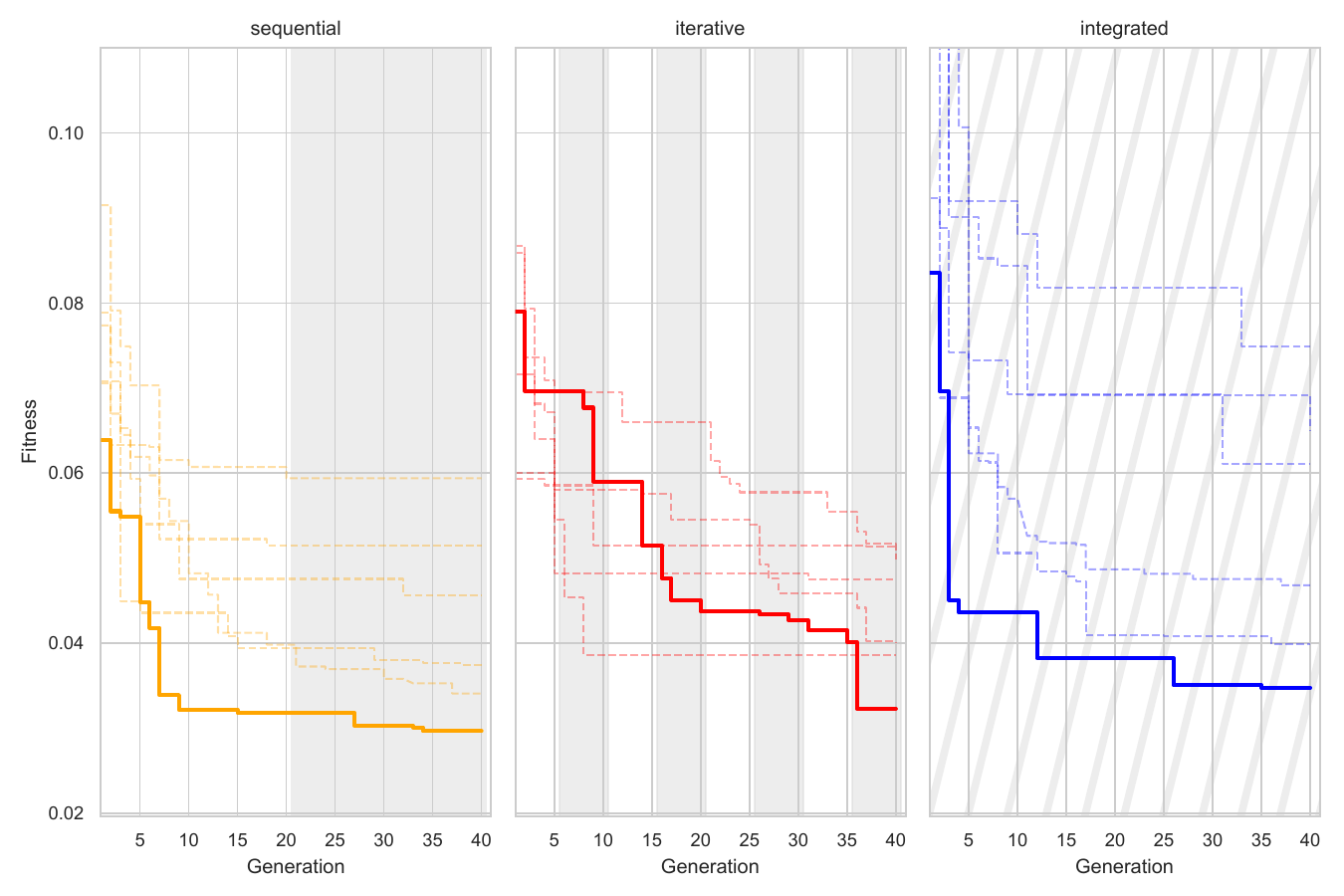}
    \caption{Fitness convergence over generations for each evolutionary coordination strategy for the IRP generated by \texttt{Qwen3-Coder:30b}. The run with the best fitness is highlighted in opaque. The vehicle routing subproblem generations are shaded in grey.}
    \label{fig:alternating_lns_IRP}
    \centering
    \includegraphics[width=1\linewidth]{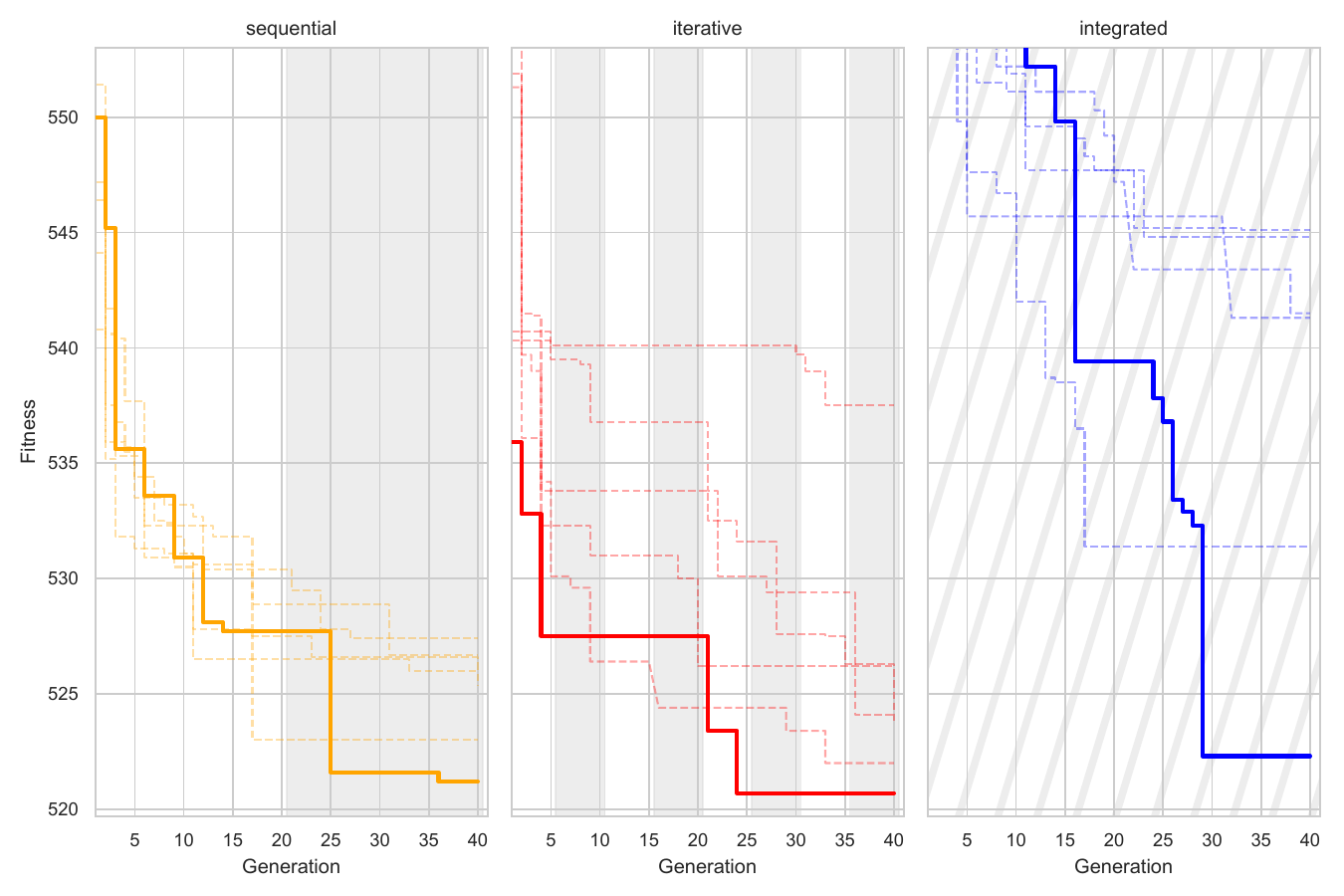}
    \caption{Fitness convergence over generations for each evolutionary coordination strategy for the MR-MUPMP generated by \texttt{Qwen3-Coder:30b}. The run with the best fitness is highlighted in opaque. The vehicle routing subproblem generations are shaded in grey.}
    \label{fig:alternating_lns_MR-MUPMP}
\end{figure}

\paragraph{Fitness Convergence.}
The fitness convergence behavior for the IRP is illustrated in Figure~\ref{fig:alternating_lns_IRP}. The experiments reveal a clear difference between the three evolutionary coordination strategies. 
\par
The sequential strategy exhibits a rapid improvement in fitness during the evolution of the heuristic for the inventory subproblem (generations 1--20) indicating the exploration of strong heuristic features.
However, after switching to the vehicle routing subproblem, the convergence rate decreases noticeably. This behavior can be attributed to two factors. 
First, the previously evolved inventory heuristic is effectively fitted to the initial vehicle routing heuristic, thereby constraining the subsequent heuristic search space and limiting further improvements. 
Second, once a high-quality heuristic has already been established through the earlier evolutionary phase, achieving additional improvements becomes inherently more difficult.
\par
In contrast, the iterative strategy demonstrates a more gradual yet consistently improving convergence trajectory. The strategy mitigates premature overfitting to heuristics and preserves adaptability throughout the evolutionary process by alternating between subproblems. 
As a result, improvements are observed across both the inventory and vehicle routing components over successive generations.
\par
For the integrated coordination strategy, the results show a heterogeneous performance pattern: while some runs achieve moderate to strong fitness values, others fail to identify competitive heuristic tuples. 
This variability suggests that the increased complexity of the joint search space hampers the consistent refinement of high-quality heuristic tuples.

The fitness convergence behavior for the MR-MUPMP is illustrated in Figure~\ref{fig:alternating_lns_MR-MUPMP}. The results show distinct convergence characteristics across the three evolutionary coordination strategies.
\par
Similar to the results for the IRP, the sequential strategy exhibits a strong improvement in fitness while evolving the first subproblem heuristic. This phase is characterized by a steep decrease in fitness values, indicating effective initial search progress. 
However, after transitioning to the second subproblem, the convergence rate decreases, albeit less pronounced than for the IRP. 
This indicates that, while the search becomes more challenging in the second phase, the  impact is weaker for the MR-MUPMP. 
The best experimental run for the sequential strategy demonstrates that improvements in this phase are still possible, but such behavior is not representative of the typical convergence trajectory.
\par
Consistent with the IRP results, the iterative strategy demonstrates a more uniform but comparatively slower convergence than the sequential strategy. 
The repeated alternation between subproblems distributes improvements across both subproblems and mitigates overfitting to a single subproblem. 
One outlier run can be observed in which the strategy fails to evolve a competitive heuristic tuple. 
Nevertheless, the best overall fitness value for the MR-MUPMP is achieved by the iterative strategy.
\par
For the integrated coordination strategy, the convergence behavior is characterized by substantial variance across runs. 
Some runs achieve significant improvements, while others show only marginal progress or remain at comparatively poor fitness levels. 
The joint evolution of both subproblems increases the difficulty of heuristic generation, which appears to hinder consistent progress and makes the discovery of high-quality heuristic tuples less reliable.
\par
Table~\ref{tab: couple_evolution_overview} summarizes the performance of the three evolutionary coordination strategies across both problem domains. 
For the IRP, the sequential strategy achieves the best minimum and mean fitness values, indicating strong overall solution quality.
The iterative strategy provides the most stable results, as reflected by the smallest median and span. 
The integrated strategy generally exhibits weaker performance, with higher average fitness values and increased variability across runs. However, it also evolves some competitive runs, indicating that it is capable of discovering high-quality solutions.
\par
For the MR-MUPMP, the sequential strategy yields the best mean performance and the smallest span, indicating a robust and consistent behavior across runs. 
The iterative strategy attains the best minimum and median fitness values, demonstrating its ability to discover a high-quality heuristic tuple. The observed variability is primarily driven by a single outlier run in which the strategy fails to identify a competitive heuristic tuple.
The integrated strategy again shows inferior performance, with the highest mean values and the largest span.

\begin{table}[hptb]
    \centering
    \caption{Fitness performance and variability across evolutionary coordination strategies for IRP and MR-MUPMP.}
    \label{tab: couple_evolution_overview}
    \begin{tabular*}{\linewidth}{@{\extracolsep{\fill}}l rrr rrr} 
    \toprule
     & \multicolumn{3}{c}{IRP} & \multicolumn{3}{c}{MR-MUPMP} \\
     \cmidrule{2-4} \cmidrule{5-7} 
     & seq. & iter. & int.  & seq. & iter. & int. \\
    \midrule
     & 0.0296 & 0.0323 & 0.0347 & 521.2 & 520.7 & 522.3 \\
     & 0.0341 & 0.0386 & 0.0398 & 523.0 & 522.0 & 531.4 \\
     & 0.0374 & 0.0400 & 0.0468 & 525.4 & 523.7 & 541.3 \\
     & 0.0456 & 0.0475 & 0.0611 & 526.0 & 524.1 & 541.5 \\
     & 0.0515 & 0.0498 & 0.0650 & 526.7 & 526.2 & 544.8 \\
     & 0.0594 & 0.0513 & 0.0748 & 527.4 & 537.5 & 545.1 \\
     \midrule
    min & \textbf{0.0296} & 0.0323 & 0.0347 & 521.2 & \textbf{520.7} & 522.3 \\
    max & 0.0594 & \textbf{0.0513} & 0.0748 & \textbf{527.4} & 537.5 & 545.1 \\
    mean & \textbf{0.0429} & 0.0433 & 0.0537 & \textbf{524.95} & 525.7 & 537.73 \\
    median & \textbf{0.0415} & 0.0437 & 0.0539 & 525.7 & \textbf{523.9} & 541.4 \\
    span & 0.0298 & \textbf{0.0191} & 0.0401 & \textbf{6.2} & 16.8 & 22.8 \\
    \bottomrule
    \end{tabular*}
\end{table}

\paragraph{Benchmarking.}
To assess the quality of the evolved heuristics, we compare the best-performing LLM-generated alternating LNS heuristics for each coupled optimization problem against established benchmark methods from the literature. This evaluation aims to position the automatically generated heuristics relative to existing approaches in terms of solution quality. We refer to our approach as \textbf{\texttt{CoupleLNS}}.
The code of the best-performing heuristics is available in our repository \href{https://github.com/tb-git-kit-research/CoupleEvo/blob/main/docs/best_heuristics.md}{(click here)}.
\par
Table~\ref{tab: IRP_benchmarking_training} reports the results for the IRP and compares the performance of \texttt{CoupleLNS} with established methods from the literature, namely the matheuristic (MH) by~\cite{archetti2017matheuristic} and the large
neighborhood and hybrid genetic search (HGS) by~\cite{ZHAO2025HGS}.
\par
The results indicate that \texttt{CoupleLNS} consistently outperforms the MH across all tested instances, achieving an average improvement of $5.61\,\%$. This demonstrates that the automatically generated heuristics can also surpass classical matheuristics in terms of solution quality.
\par
In comparison to HGS, which represents a state-of-the-art approach for the IRP, \texttt{CoupleLNS} achieves slightly inferior results. The average gap of $2.96\,\%$ suggests that while the LLM-generated heuristics do not outperform highly specialized methods, they still achieve competitive performance without relying on extensive problem-specific engineering.

\begin{table}[hptb]
    \caption{Benchmark comparison of solution quality on IRP training instances. Gaps are reported in \%.}
    \label{tab: IRP_benchmarking_training}
    \begin{tabular*}{\linewidth}{@{\extracolsep{\fill}}l rrrrr} 
    \toprule
    N Q T & MH & HGS & \texttt{CoupleLNS} & MH Gap & HGS Gap \\
    \midrule
    25 18 6  & 3426.5 & 3211.4 & 3242.6 & -5.37 & 0.97 \\
    25 18 9  & 4491.2 & 4193.0 & 4361.6 & -2.88 & 4.02 \\
    25 18 12 & 8721.1 & 7672.5 & 7971.5 & -8.59 & 3.90 \\
    \midrule
    Mean  & 5546.27 & 5025.63 & 5191.91 & -5.61 & 2.96 \\
    \bottomrule
    \end{tabular*}
\end{table}

Table~\ref{tab: mr-mupmp_benchmark_training} reports the benchmarking results for the MR-MUPMP and compares \texttt{CoupleLNS} against the state-of-the-art step-based sequential solution approach (SBS) proposed by~\cite{BOMER2026107359sorting} on the same set of problem instances generated by the seeds zero to nine. The SBS relies on an optimal A* search to minimize the number of reshuffling moves and tie-breaks on loaded move distance. However, due to the complexity of the move search subproblem, this optimal search fails to find feasible solutions within the imposed time limit of 600 seconds for some instances.
\par
The results indicate that, for most instances for which a solution is obtained, SBS achieves superior solution quality compared to \texttt{CoupleLNS} (average gap of $5.04\,\%$). However, the approach fails to solve some instances (seeds two and four) within the time limit. In contrast, \texttt{CoupleLNS} successfully solves all instances, albeit with slightly inferior average solution quality.
The performance difference is partly caused by the number of moves found in the move search subproblem.

\begin{table}[hptb]
    \caption{Benchmark comparison of solution quality on MR-MUPMP training instances. Each row shows the result for a problem instance generated with a certain seed. Our approach is listed as CoupleLNS. Gaps are reported in \%.}
    \label{tab: mr-mupmp_benchmark_training}
    \begin{tabular*}{\linewidth}{@{\extracolsep{\fill}}c|rrrrr}
    \toprule
    Seed & SBS & SBS* & \texttt{CoupleLNS} & SBS Gap& SBS* Gap \\
    \midrule
    0 & 494 & 564 & 541 & 9.51 & -4.08 \\
    1 & 467 & 557 & 477 & 2.14 & -14.36 \\
    2 & - & 574 & 572 & - & -0.35 \\
    3 & 595 & 627 & 600 & 0.84 & -4.31 \\
    4 & - & 608 & 567 & - & -6.74 \\
    5 & 420 & 434 & 431 & 2.62 & -0.69 \\
    6 & 419 & 467 & 420 & 0.24 & -10.06 \\
    7 & 508 & 615 & 578 & 13.78 & -6.02 \\
    8 & 510 & 571 & 568 & 11.37 & -0.53 \\
    9 & 454 & 467 & 453 & -0.22 & -3 \\
    \midrule
    Mean & 483.38 & 548.40 & 520.70 & 5.04 & -5.01 \\
    \bottomrule
    \end{tabular*}
\end{table}

\par
To provide a fair comparison, we additionally consider an adapted variant, denoted as SBS*, in which the optimal A* is replaced by an A* search guided by an LLM-generated heuristic as proposed in~\cite{bomer2026algorithmic}. This modification enables the approach to solve all instances within the time limit, albeit without guaranteeing optimality with respect to the number of moves.
\texttt{CoupleLNS} and SBS* use the same A* guided by an LLM-generated heuristic.
\par
In comparison to SBS*, \texttt{CoupleLNS} achieves competitive performance, with an average improvement of $5.01\,\%$. While SBS* benefits from the LLM-guided search to ensure feasibility across all instances, the results indicate that \texttt{CoupleLNS} is able to produce solutions of superior quality. Notably, \texttt{CoupleLNS} outperforms SBS* on all instances.

\section{Conclusion}
\label{sec: Conclusion}
In this work, we introduced \textit{CoupleEvo}, the first LLM-driven framework for automated heuristic design in coupled optimization problems, extending existing approaches beyond single-problem settings. We proposed three evolutionary coordination strategies (sequential, iterative, and integrated) that govern how heuristics are generated and refined.
\par
Our experimental results on two representative coupled optimization problems, the IRP and the MR-MUPMP, demonstrate that the choice of coordination strategy has a significant impact on convergence behavior, robustness, and final heuristic quality. 
In particular, decomposition-based strategies (sequential and iterative) consistently yield more reliable performance than the integrated evolution strategy.
\par
The sequential evolution strategy is characterized by rapid initial fitness convergence during the heuristic evolution of the first subproblem. 
However, this early convergence often limits further improvements after switching to the second subproblem, as the search becomes constrained by the previously evolved heuristic that is highly fitted to the initial seeding heuristic of the other subproblem. 
The iterative evolution strategy exhibits a more uniform convergence behavior. By repeatedly alternating between subproblems, it maintains flexibility throughout the search process. This enables continuous improvements across both subproblems.
In contrast, the integrated evolution strategy exhibits higher variance and struggles with the increased complexity of jointly evolving multiple heuristics. Yet, it remains capable of producing competitive solutions in individual runs.
\par
The benchmarking results show that the evolved \texttt{CoupleLNS} approach is competitive with, and in some cases superior to, established methods from the literature. 
For the IRP, \texttt{CoupleLNS} outperforms classical matheuristics and approaches the performance of state-of-the-art hybrid metaheuristics. 
\texttt{CoupleLNS} demonstrates a favorable trade-off between solution quality and robustness for the MR-MUPMP, consistently solving all instances while remaining competitive with step-based approaches.
\par
Overall, these findings highlight the potential of LLM-driven automated heuristic design for coupled optimization problems. Rather than replacing classical methods, the results suggest that LLM-based approaches can complement existing techniques like LNS by evolving highly fitted problem-specific heuristic components.
\par
Several avenues for future research emerge from this work. 
First, adapting the framework to other coupled optimization problems with more than two subproblems.
Second, improving the stability and scalability of the integrated evolution strategy remains an important challenge, particularly for problems with strong inter-dependencies or a larger number of subproblems.
Finally, developing adaptive coordination strategies that dynamically switch between optimization problems may further enhance performance.

\begin{acks}
This research was funded by the Ministry of Science, Research and Arts of the Federal State of Baden-Württemberg - InnovationCampus Future Mobility - funding code BUP59 FitLLM and the European Union - NextGenerationEU - funding code 13IK032I.
\end{acks}


\bibliographystyle{ACM-Reference-Format}
\bibliography{sources}
\end{document}